\documentclass[11pt]{pdfpages}

\usepackage{pdfpages}

\begin{document}

\title[Article Title]{PolyCL: Contrastive Learning for Polymer Representation Learning via Explicit and Implicit Augmentations
}

\author[1]{\fnm{Jiajun} \sur{Zhou}}

\author[1]{\fnm{Yijie} \sur{Yang}}

\author[1,2]{\fnm{Austin M.} \sur{Mroz}}

\author*[1]{\fnm{Kim E.}\sur{Jelfs}}\email{k.jelfs@imperial.ac.uk}

\affil[1]{\orgdiv{Department of Chemistry}, \orgname{Imperial College London}, \orgaddress{White City Campus, London, W12 0BZ, \country{United Kingdom}}}

\affil[2]{\orgdiv{I-X Centre for AI in Science}, \orgname{Imperial College London}, \orgaddress{White City Campus, London, W12 0BZ, \country{United Kingdom}}}

\abstract{Polymers play a crucial role in a wide array of applications due to their diverse and tunable properties. Establishing the relationship between polymer representations and their properties is crucial to the computational design and screening of potential polymers via machine learning. The quality of the representation significantly influences the effectiveness of these computational methods. Here, we present a self-supervised contrastive learning paradigm, PolyCL, for learning high-quality polymer representation without the need for labels. Our model combines explicit and implicit augmentation strategies for improved learning performance. The results demonstrate that our model achieves either better, or highly competitive, performances on transfer learning tasks as a feature extractor without an overcomplicated training strategy or hyperparameter optimisation. Further enhancing the efficacy of our model, we conducted extensive analyses on various augmentation combinations used in contrastive learning. This led to identifying the most effective combination to maximise PolyCL's performance.}

\keywords{Polymer Science, Machine Learning, Self-supervised Learning}

\maketitle

\section{Introduction}
\label{sec:introduction}

Polymers, with their remarkable diversity and extensive adaptability, have emerged as a key material class across various applications,\cite{sha2021machine} including medicine and medical devices,\cite{maitz2015applications} agriculture,\cite{puoci2008polymer} solar cells,\cite{li2012polymer} and electronics.\cite{jaiswal2006polymer}
Polymers are made from combinations of small, organic molecule-based monomeric building blocks and thus there is an enormous chemical space to be explored. The complexity of polymers can also be reflected in the extended polymer material space, including the variety of processing and synthetic conditions for the production of polymer products that can vary their performance.\cite{matyjaszewski2005macromolecular, sada2018functional, binder2007click} The effective exploration of the extensive chemical space of polymers is a major challenge in the discovery of functional polymers for target applications. Indeed, this space is far too large to feasibly explore with conventional trial-and-improvement experimental approaches alone. The integration of computational modelling and machine learning has significantly accelerated this process, enabling the rapid identification of promising candidates.\cite{chen2021polymer, mannodi2016machine} However, there exist many challenges to training robust ML models for polymer property prediction, including limited high-quality data,\cite{martin2023emerging} scarcity of data in a specific property space,\cite{kuenneth2021polymer} and highly diverse polymer representations.\cite{mannodi2016machine, zeng2018graph, lin2019bigsmiles, doan2020machine} Indeed, polymer representations pose a challenge for many reasons, including difficulties describing repeating structures built from monomeric units, and the lack of representations that incorporate macroscopic packing.\cite{martin2023emerging, lin2019bigsmiles} Thus, designing new polymer representations is an active area of research.

The design of polymer representations, which refers to the machine-readable way that the molecular features of polymers are encoded, is critical for the performance of property prediction models. Conventional methods for creating machine-readable polymer representations involve creating handcrafted fingerprints, where molecular structural information features are depicted through manually designed descriptors\cite{kuenneth2023polybert}. Indeed, several types of handcrafted fingerprints\cite{tao2021machine, zhu2020machine, wu2019machine} and refined fingerprint strategies\cite{mannodi2016machine, doan2020machine} within this category have found success in the polymer literature. While these methods have found success, handcrafted fingerprints are often designed using the expert's chemical intuition and heuristic principles. In addition to the potential to introduce bias, these methods are fairly labour-intensive and time-consuming.

\begin{figure}
    \centering
	\includegraphics[width=0.6\linewidth]{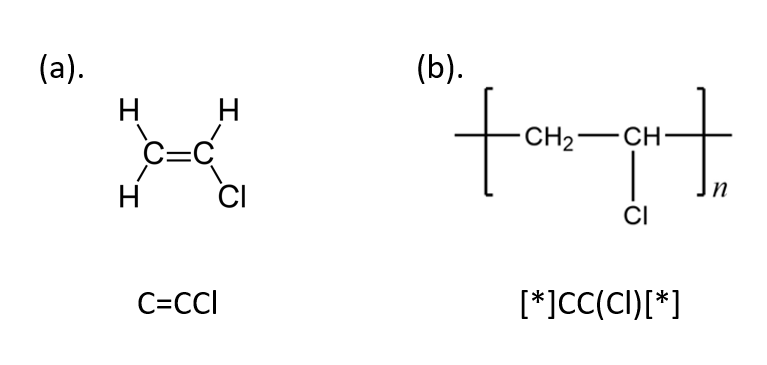}
	\caption{Example of (a) SMILES of vinyl chloride and (b) polymer-SMILES of polyvinyl chloride, along with the corresponding chemical structures.}
	\label{fig:polymersmile}
\end{figure}

Deep neural networks have been increasingly used to automatically extract dense molecular representations from polymers.\cite{zhou2020graph} This approach leverages the power of deep learning to alleviate the aforementioned challenges associated with manual feature extraction. Polymers can be abstracted into molecular graphs.\cite{zeng2018graph, queen2023polymer, park2022prediction} Alternatively, molecules can be converted to one-dimensional sequence representations, such as SMILES.\cite{weininger1988smiles} Here, polymer-SMILES representations are used that include brackets with a special character ``[*]" to represent connection points between monomers, to reproduce the repeating nature of these materials (example shown in Fig. \ref{fig:polymersmile}). For example, a Long Short-Term Memory (LSTM) model was trained on polymers represented as SMILES strings for property prediction.\cite{chen2021predicting, mannodi2016machine} In addition, BigSMILES was designed to extend SMILES for the representation of the stochasticity of polymer molecules by introducing additional notations.\cite{lin2019bigsmiles}

Machine learning-based predictive models are typically trained in a supervised fashion and act as automatic feature extractors. While this supervised training pattern is beneficial for specific downstream tasks, it may lead to learnt representations exhibiting domain-dependent characteristics and suffering from limited generalisability to other tasks.\cite{irwin2022chemformer, cui2018large, he2019rethinking, tendle2021study} Further, supervised learning methods rely on labelled data of both high quantity and high quality. Within (polymer) chemistry, acquiring high-quality, labelled data is resource-intensive. The scarcity of labelled data in chemistry\cite{sun2019infograph} may lead to overfitting and, therefore, impair the model's generalisability to other data in the target domain.\cite{wang2022molecular} The limitations of supervised learning directly motivates self-supervised learning for chemical property prediction.

Self-supervised models learn from the inherent structure of the data, without the requirement of data labelling.\cite{balestriero2023cookbook} In polymer science, the value of creating a universal representation that is a target-agnostic feature of self-supervised learning has already been observed.\cite{xu2023transpolymer}  Initial demonstrations of self-supervised learning in polymer science have largely focused upon transformer architectures.\cite{vaswani2017attention} Masked language modeling,\cite{devlin2018bert} where random tokens are obscured from the input to be predicted by the transformer, served as the training strategy to guide the pre-training of transformers. This training strategy was proven effective in Transpolymer and polyBERT for the production of machine-learnt polymer representations using transformer architecture.\cite{xu2023transpolymer, kuenneth2023polybert} However, these works did not directly assess the effectiveness of the representation learnt by their pre-trained strategies and only inferred the quality of the learned representation from the performance of the downstream tasks using the model. Yet, inferring representation performance from the model performance is especially challenging when the model includes complex ML techniques, such as data-augmentation (including non-canonical SMILES strings to increase the size of the dataset),\cite{xu2023transpolymer} and multi-task learning (training downstream tasks on all datasets simultaneously).\cite{kuenneth2023polybert}

Contrastive learning is among the most competitive forms of self-supervised learning that learn meaningful representations from comparing and contrasting data.\cite{tian2020makes} The idea of contrastive learning is to pull together similar data samples and separate dissimilar samples in the representation space.\cite{yang2022mutual} This idea has been demonstrated to achieve successful representation learning in molecular systems.\cite{you2020graph, yin2022autogcl, cao2023moformer} In addition, contrastive learning is capable of incorporating extra modality to form modality pairs such as structures and text description,\cite{liu2023multi} SMILES and IUPAC names,\cite{guo2021multilingual} SMILES and the molecular graph,\cite{pinheiro2022smiclr} into the molecular representation via multi-modal alignment. However, contrastive learning is yet, to the best of our knowledge, to be applied to polymer science.

\begin{figure}
    \centering
	\includegraphics[width=\linewidth]{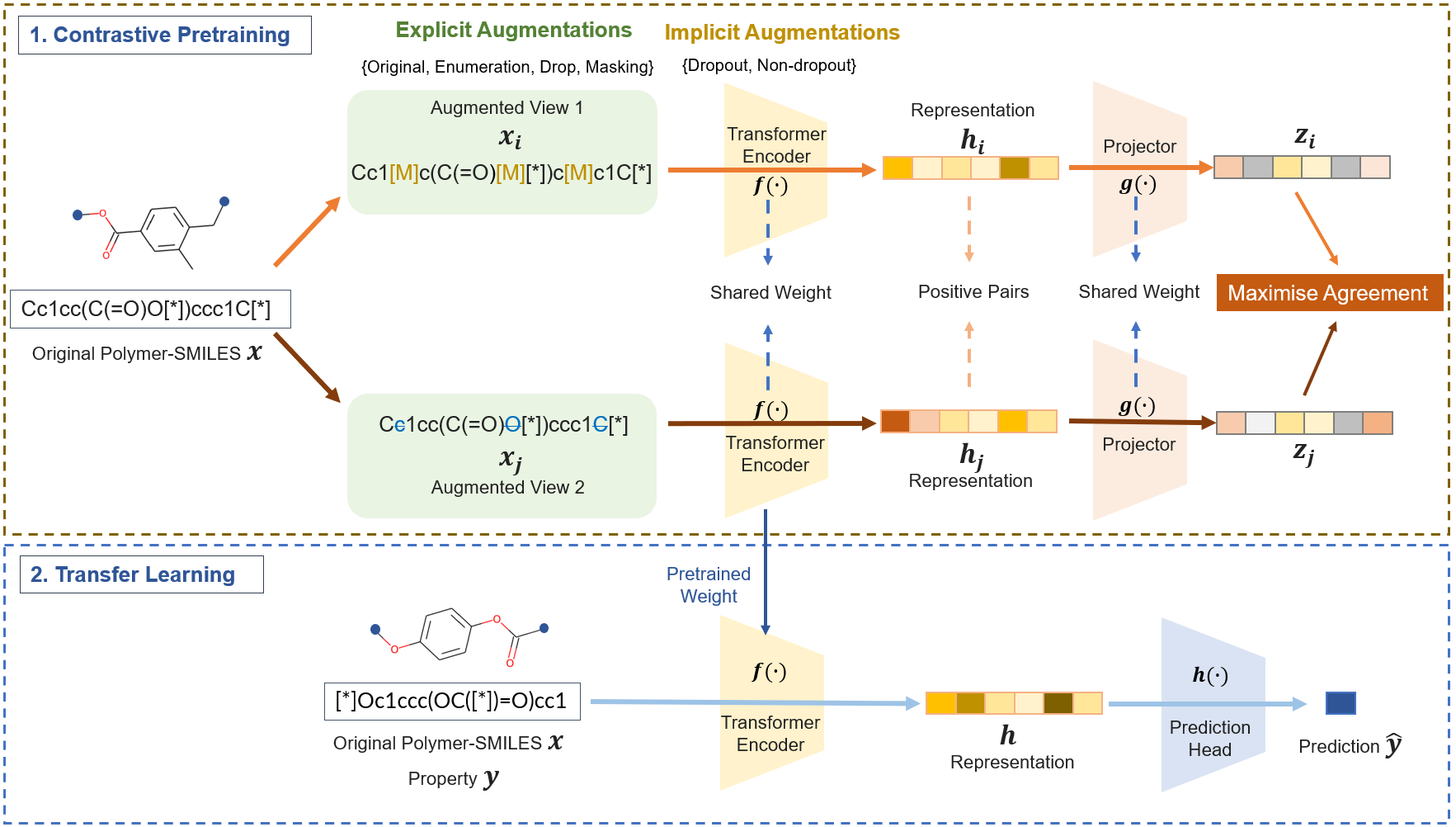}
	\caption{A schematic illustration of the PolyCL pipeline. (1) Polymer contrastive representation learning with different augmentation strategies for constructing effective positive pairs. The agreement of positive pairs projected to their latent representations is maximised by the loss function of contrastive learning. Masking and Drop in augmented views 1 and 2 are shown as sample explicit augmentations for the input original polymer-SMILES. (2) Transfer learning by leveraging the acquired polymer representation to apply in the prediction of downstream tasks.}
	\label{fig:cl_schematic}
\end{figure}

As illustrated in Figure \ref{fig:cl_schematic}, an efficient approach to contrastive learning entails the formation of positive pairs by creating two distinct representations of the same, original polymer-SMILES molecule (here, termed \textit{anchor molecule}) through data augmentation. This process is critical as it enables the learning model to recognise and reinforce the essential features of the molecule by comparing these different views. Concurrently, the anchor molecule and other molecules in the current (with their respective positive pair) are automatically considered negative pairs.\cite{chen2020simple} Thus, the construction of positive pairs is exceptionally important because their formation directly impacts the identification of negative pairs, which are imperative to helping the algorithm understand the relationship between different positive pairs and, ultimately, better map out the representation space.

In chemistry, common approaches to augmentation are explicit -- allowing observable modifications to the representation structure (\textit{e.g.} removing a token from a SMILES string). Typical explicit augmentation modes for molecular graphs include node dropping, edge masking/perturbation, attribute masking and subgraph extraction.\cite{you2020graph,pinheiro2022smiclr} Implementation of explicit augmentation methods for SMILES representations remain limited and under-explored.\cite{pinheiro2022smiclr}  In addition, augmentation can also be implemented in an implicit fashion, where different perturbations to the embedding are implemented during the training process (\textit{e.g.} natural dropout).\cite{gao2021simcse, xia2022simgrace} Despite the demonstrated effectiveness of implicit augmentation, this approach also remains an area of limited attention. Furthermore, there is a need to understand the effects arising from the heterogeneous combination of both types of augmentation strategies (\textit{i.e.} implicit and explicit).

Here, we present PolyCL, a contrastive learning framework for polymer representation learning for improved predictive performance. To construct effective positive pairs, we also proposed a novel combinatorial augmentation strategy to include both explicit and implicit augmentations. Our results show that PolyCL outperforms other supervised and pre-trained models under the lightweight and flexible transfer learning setting where the fine-tuning of PolyCL is not required. Here, we emphasise that this construction eliminates the need to fine-tune an entire model (\textit{e.g.} pre-trained model + prediction head), instead PolyCL may be independently implemented as a feature extractor for polymer property prediction tasks. We find that the learnt representation from our contrastive learning strategy has improved quality, and show how our polymer representation can be used for a variety of downstream tasks via simple transfer learning. The dataset and model are available at: \url{http://github.com/JiajunZhou96/PolyCL}.

\section{Methods}
\label{sec:methods}
\subsection{Dataset}
 We randomly selected 1 million polymers from the unsupervised polymer dataset curated by Xu {\em {et al.}}\cite{xu2023transpolymer} to use as the pre-train dataset for contrastive learning. Datasets for downstream regression tasks were sourced from data by Xu {\em {et al.}}\cite{xu2023transpolymer} to benchmark against other models. Specifically, we focused on homopolymer datasets, where the inputs are comprised only of the SMILES strings of the monomers. For extension of the approach to copolymers or multi-component polyelectrolyte systems in the future, extra descriptors can be easily concatenated with the polymer representations produced from our model to collaboratively encode additional information. We used seven different property datasets covering a wide range including band gap (both chain (Egc) and bulk (Egb)), electron affinity (Eea), ionisation energy (Ei), Density Functional Theory (DFT)-calculated dielectric constant (EPS), crystallisation tendency (Xc), and refractive index (Nc). We did not use any data augmentation strategy to boost our downstream datasets. All datasets were originally calculated by DFT calculations.\cite{kuenneth2021polymer}

\subsection{Polymer Encoding}

Polymers are often linearly concatenated by the repeating units of monomers, exhibiting inherently sequential structures.\cite{rudin2012elements} Therefore, there are advantages to representing a polymer as a sequence-based molecular representation. SMILES strings\cite{weininger1988smiles} are commonly employed for depicting individual monomers within polymers. Different to the representation of small molecules, polymers necessitate the explicit indication of connecting points between monomers. As we start the training process with the pre-trained checkpoint of polyBERT,\cite{kuenneth2023polybert} we maintained the use of polymer-SMILES to make full use of the model. In comparison, polymer-SMILES extends the traditional SMILES representation by marking connecting points with the special token ``[*]'', following the standard syntactic rules of the SMILES format. Subsequently, the input polymer-SMILES were encoded by the pre-trained polyBERT model with the corresponding tokeniser,\cite{kuenneth2023polybert} which is a variation of the Deberta-v2 \cite{he2020deberta} language model with a transformer architecture.\cite{vaswani2017attention}

\subsection{Contrastive Representation Learning Objective}
To effectively guide the training of the model to the intended objective, we applied the normalised temperature-scaled cross-entropy (NT-Xent) loss.\cite{chen2020simple} In a batch consisting of $2N$ semantically similar views derived from $N$ samples, for each positive pair $(i, j)$, the remaining $2(N-1)$ samples in the batch are implicitly considered as negative examples. Therefore, the NT-Xent loss for a positive pair $(i, j)$ is described by

\begin{equation}
    \mathcal{L}_{i, j}=-\log \frac{\exp \left(\operatorname{sim}\left(z_i, z_j\right) / \tau\right)}{\sum_{k=1}^{2 N} \mathbbm{1}_{\{k \neq i\}} \exp \left(\operatorname{sim}\left(z_i, z_k\right) / \tau\right)}
\label{eq:contrastive_learning}
\end{equation}

\noindent where $z_i$, $z_j$ are the representations of two positive data samples, $\operatorname{sim}(u, v)$ denotes the cosine similarity $\frac{{u}^{T}v}{{\lVert u\rVert}{\lVert v\rVert}}$, $\tau$ is the temperature parameter, which is empirically set to 0.05. An indicator function $\mathbbm{1}_{\{k \neq i\}}$ is used to skip the case where both $k$ and $i$ refer to the same sample.

Here, we used the pre-trained PolyBERT\cite{kuenneth2023polybert} as our encoder, $\bm{f(\cdot)}$, and maintained the default settings for all hyperparameters in the transformer architecture. The projector $\bm{g(\cdot)}$ is a two-layer MLP that maps the pooled 600-dimensional representation $\bm{h}$ to a 128-dimensional latent vector $\bm{z}$ for similarity evaluations. During the contrastive pre-training, we enabled mixed precision training. AdamW\cite{loshchilov2017decoupled} was used as the optimiser with a learning rate of 1e-5 to minimise the NT-Xent loss. A gradient clipping mechanism was employed with a max grad norm set to 1.0. We trained the model for 10 epochs in total.

\subsection{Constructing Augmentations}
Contrastive learning can be enhanced by the use of effective data augmentation modes, a benefit observed across various data modalities.\cite{chen2020simple, you2020graph, pinheiro2022smiclr} The challenge for contrastive learning is the construction of effective positive pairs. This can be achieved by applying augmentation strategies to create different views of the same polymer molecules, which should subtly alter the attributes of the polymer representations. We aim to create differences in two vectors of polymer representations $\bm{h_i}$ and $\bm{h_j}$, while preserving the key semantic information referring to the original polymer molecule $\bm{x}$. In this case, the use of the original molecule can be considered as the baseline.

Augmentations can be empirically categorised into two modes; ``explicit'' and ``implicit''. Explicit augmentations are direct and observable modifications to the input data. As shown in Fig. \ref{fig:cl_schematic}, explicit augmentations include enumeration, token masking (Masking) and token drop (Drop). Enumeration creates one random non-canonical SMILES string of the polymer. Drop deletes 10\% of tokens in the SMILES string. Masking substitutes 10\% of tokens in the SMILES string with a special token. Therefore, the same molecule can be transformed into two different SMILES strings to construct an effective positive pair. 

Beyond explicit augmentations, the subtle modifications in how the input data is represented in the intermediate layers within the model are referred to as implicit augmentations, as shown in Fig. \ref{fig:cl_schematic}. Following the work of SimCSE,\cite{chen2020simple} we used the inherent dropout module inside our transformer encoder to create differences in molecular embedding for the same input. With implicit augmentations enabled, the dropout ratio for hidden layers and attention probabilities in the configuration of the transformer encoder is 0.1; when disabled, both values are set to 0. In addition, we have also combined both explicit and implicit augmentations for the construction of positive pairs to study the cooperative effect of augmentation strategies.

\subsection{Transfer Learning}
\label{sec:transfer_learning}
We used transfer learning to evaluate the quality of learnt representations. We fine tune the prediction head and leave the pre-trained model unchanged during transfer learning. In the implementation of this approach, all trainable parameters in the pre-trained model were frozen and gradients were turned off before the training of transfer learning models of downstream tasks.  

The experimental setup employs an MLP regressor featuring a single hidden layer and ReLU activation, integrated with the PolyCL feature extractor. ``[CLS]'' pooling serves as the readout function, extracting a 600-dimensional polymer representation. Specifically, this approach transforms token-level embeddings for each polymer sequence into a comprehensive sentence-level embedding, wherein sequence information is encapsulated by the appended ``[CLS]'' token. The hidden size within the MLP is consistent with the input size for all pre-trained models (including the benchmarking study of PolyBERT and Transpolymer). A dropout ratio of 0.1 is applied. polymer-SMILES strings are encoded by the tokeniser of PolyBERT \cite{kuenneth2023polybert}. An $\textit{l}_{2}$ loss function is implemented for regression tasks. During the regression phase, AdamW\cite{loshchilov2017decoupled} was used as the optimiser with a learning rate of 0.001 and no weight decay. For each downstream dataset, a 5-fold cross-validation strategy is employed, accompanied by a 500-epoch training protocol. An early-stopping monitor is activated after 50 epochs, with a patience setting of 50 epochs. The performance on the unseen validation datasets is evaluated using the root-mean-square error (RMSE) and the coefficient of determination ($\textit{R}^{2}$). To show the general expressiveness of the learnt representation, all hyperparameters for the transfer learning performed on PolyCL are set by simple heuristics and not tuned specifically by a validation process.

\subsection{Alignment and Uniformity}
The quality of the learned representation can be alternatively evaluated by the quantitative metrics of alignment and uniformity introduced by Wang and Isola.\cite{wang2020understanding} Alignment refers to the distance between known positive pairs $(x, x^{+}) \sim p_{\text {pos}}$, as shown in Equation \ref{eq:alignment}. A lower alignment value between positive pairs indicates improved feature similarity:

\begin{equation}
    \ell_{\text {align }} \triangleq \underset{\left(x, x^{+}\right) \sim p_{\text {pos}}}{\mathbb{E}}\left\|f(x)-f\left(x^{+}\right)\right\|^2
\label{eq:alignment}
\end{equation}

\hspace{-0.65cm} where $x$ is a polymer-SMILES, $x^{+}$ is a known positive view to $x$. $f(x)$ is a neural encoder to transfer polymer-SMILES to a representation. ${\mathbb{E}}$ is the expectation.

Uniformity is a measure of the distribution of learnt representations in the unit hypersphere; this is defined by the log of the mean Gaussian potential between each embedding pair ${{x, y \stackrel{i .i. d .}{\sim} p_{\text {data}}}}$, where each variable in the pair is an independent and identically distributed random variable, as shown in Equation \ref{eq:uniformity}. A lower uniformity indicates the learnt embedding distribution is capable of preserving maximal information:

\begin{equation}
    \ell_{\text {uniform }} \triangleq \log \underset{\substack{x, y \stackrel{i . i .d .}{\sim} p_{\text {data }}}}{\mathbb{E}} e^{-2\|f(x)-f(y)\|^2}
\label{eq:uniformity}
\end{equation}

To effectively evaluate the alignment and uniformity, a dataset distribution that has never been seen by any of the pre-trained models needed to be constructed to ensure a fair cross-model comparison. Here, we randomly sampled 60,000 polymers from the excluded development dataset of polyBERT\cite{kuenneth2023polybert} for evaluation. Each polymer was augmented once by the SMILES enumeration method to create a positive pair, thereby preserving semantics.

\subsection{Benchmarking Other Models}
The implementation details of all supervised learning models are shown in Supplementary Information Section S6, including random forest, XGBoost, neural networks, GCN and GIN. For all pre-trained models, we froze all parameters and consolidated the pooling method as ``[CLS]'' pooling. We also used a simple MLP regressor as shown in Section \ref{sec:transfer_learning} with only adaptation on the input layer size to fit different sizes of input representations from different pre-trained models.

\section{Results}
\label{sec:results}

\subsection{Polymer Contrastive Learning}

The PolyCL architecture for obtaining a machine-learned polymer representation is shown in Fig. \ref{fig:cl_schematic}. In the pre-training phase, the repeating units of polymers were encoded to polymer-SMILES, $\bm{x}$.\cite{kuenneth2023polybert} Then, we converted each original $\bm{x}$ into two views $\bm{x_i}$ and $\bm{x_j}$, {\em{i.e.}} positive pairs in two branches of the model. All views are processed by a transformer encoder $\bm{f(\cdot)}$ to obtain the contextualised embedding. Here, we used the pre-trained polyBERT model\cite{kuenneth2023polybert} as the encoder to obtain a more effective prior than random initialisation, for subsequent fine-tuning by our PolyCL framework. Then, we applied [CLS] pooling, which generates compressed representations of the polymer-SMILES\cite{reimers2019sentence,chen2020simple} on the contextualised embedding to obtain the polymer representation $\bm{h_i}$ and $\bm{h_j}$. The projector was introduced by SimCLR,\cite{chen2020simple} which inspired the architecture of PolyCL. Here, these pooled representations $\bm{h_i}$ and $\bm{h_j}$ are further projected as $\bm{z_i}$ and $\bm{z_j}$ using a projector $\bm{g(\cdot)}$ into a latent space. Additionally, any pairs in which the source instances within each pair originate from different original polymer molecules are considered negative pairs. The objective function of contrastive learning is the normalised temperature-scaled cross-entropy (NT-Xent) loss, aiming to develop machine-learned representations by attracting positive pairs while distancing negative pairs in the latent space.\cite{hadsell2006dimensionality}

In the transfer learning phase, we extracted the representation using the pre-trained model and then used a prediction head to predict any property of interest. The pre-trained transformer encoder is employed to encode polymers to their representations. Here, we demonstrate how the prediction process using a simple prediction head $\bm{h(\cdot)}$, constructed with two-layered multi-layer perceptrons (MLP) with random initialisation, can be used to train the mapping from polymer representations to properties $\bm{\hat{y}}$. However, the transfer learning process is flexible in selecting predictive models that best serve the requirements of downstream tasks.

\subsection{Transfer Learning Results}
\label{sec:transfer_learning_results}

The primary objective of our study is to create an effective and expressive machine-learnt representation for polymers. Transfer learning is employed to assess the utility of knowledge extracted from a pre-trained model. To evaluate the expressiveness of the representation, polymer representations produced by PolyCL are directly adopted without any task-specific refinement. In practice, we achieved our objective by fine-tuning only the prediction head, while keeping all parameters of the pre-trained model frozen. 

\begin{table}[t]
\caption{The average $R^2$ values on the unseen validation datasets with five-fold cross-validation. Seven polymer property datasets were used for predictive benchmarking: band gap (both chain (Egc) and bulk (Egb)), electron affinity (Eea), ionisation energy (Ei), DFT-calculated dielectric constant (EPS), crystallisation tendency (Xc), and refractive index (Nc). $\rm{RF_{ECFP} }$, $\rm{XGB_{ECFP}}$, $\rm{NN_{ECFP}}$, $\rm{GP_{PG}}$, $\rm{NN_{PG}}$, GCN and GIN are supervised models. TransPolymer, PolyBERT and PolyCL are self-supervised models. `\#Params' indicates the number of parameters. The numbers in \textbf{bold} indicate the best results for a given property.}
\label{tab:overall_performance_R2}
\small %
\setlength{\tabcolsep}{2pt} %
\renewcommand{\arraystretch}{1} %
\begin{tabular*}{\textwidth}{cc | ccccccc | c}
\toprule
\multicolumn{2}{c}{Model information} & \multicolumn{8}{c}{Datasets} \\
\cmidrule{1-2}\cmidrule{3-10}
Model & \#Params & Eea & Egb & Egc & Ei & EPS & Nc & Xc & Avg.$R^2$\\
\midrule
$\rm{RF_{ECFP} }$ & - & 0.8401 & 0.8643 & 0.8704 & 0.7421 & 0.6840 & 0.7540 & 0.4345 & 0.7413\\
$\rm{XGB_{ECFP}}$ & - & 0.8350 & 0.8568 & 0.8679 & 0.7221 & 0.6728 & 0.7574 & 0.3842 & 0.7280\\
$\rm{NN_{ECFP}}$ & 264K & 0.8543 & 0.8708 & 0.8838 & 0.7562 & 0.7473 & 0.8066 & 0.3975 & 0.7595\\
$\rm{GP_{PG}}$ \footnotemark[1] & - & 0.90 & \textbf{0.91} & \textbf{0.90} & 0.77 & 0.68 & 0.79 & $\rm{<0}$ & $\rm{<0.71}$\\
$\rm{NN_{PG}}$ \footnotemark[2] & - & 0.87 & 0.90 & 0.89 & 0.74 & 0.71 & 0.78 & $\rm{<0}$ & $\rm{<0.70}$\\
GCN & 70K & 0.8544 & 0.8043 & 0.7988 & 0.6646 & 0.7404 & 0.5238 & 0.3316 & 0.6739\\
GIN & 218K & 0.8829 & 0.8350 & 0.8181 & 0.7841 & 0.6925 & 0.6317 & 0.3902 & 0.7192\\
\midrule
TransPolymer \cite{xu2023transpolymer} & 82.1M & 0.8943 & 0.8961 & 0.8756 & 0.7919 & 0.7568 & 0.8109 & \textbf{0.4552} & 0.7830\\
PolyBERT \cite{kuenneth2023polybert} & 25.2M & 0.9065 & 0.8830 & 0.8783 & 0.7670 & 0.7694 & 0.8017 & 0.4367 & 0.7775\\
$\rm{PolyCL}$ & 25.2M & \textbf{0.9071} & 0.8884 & 0.8832 & \textbf{0.8112} & \textbf{0.7876} & \textbf{0.8460} & 0.4043 & \textbf{0.7897}\\
\bottomrule
\end{tabular*}
\footnotetext[1,2]{The $R^2$ values of these two lines are directly taken from the single-task learning experiments of Kuenneth {\em {et al.}}.\cite{kuenneth2021polymer}}
\end{table}

There are two key advantages to this strategy. Firstly, this approach ensures that the representation is independent of the further fine-tuning of the underlying pre-trained model during the transfer learning, allowing for a fairer evaluation of the representation's quality. Secondly, this approach aligns with common real-world applications better; typically, only the polymer representation is incorporated into subsequent models, instead of the specialised fine-tuning of the pre-trained model with these later models. PolyCL should serve simply as a flexible representation generator and, therefore, requires no extra computational resources to fine-tune pre-trained parameters during usage. To explore how our model performs, we compared PolyCL with supervised models including random forest (RF), XGBoost (XGB), and neural networks (NN), each trained on either ECFP fingerprints\cite{rogers2010extended} or the domain-specific Polymer Genome (PG) fingerprints.\cite{kuenneth2021polymer} We also implemented cross-modal comparison between the above fingerprints and graph representations encoded by graph convolutional networks (GCN)\cite{kipf2016semi} and graph isomorphic networks (GIN).\cite{xu2018powerful} Finally, we compared with other machine-learnt representations via self-supervised learning strategies: PolyBERT\cite{kuenneth2023polybert} and Transpolymer.\cite{xu2023transpolymer}

The results of our transfer learning are shown in Table \ref{tab:overall_performance_R2}.  We conducted our transfer learning on seven different datasets sourced from Xu {\em {et al.}},\cite{xu2023transpolymer} including band gap (both chain (Egc) and bulk (Egb)), electron affinity (Eea), ionisation energy (Ei), DFT-calculated dielectric constant (EPS), crystallisation tendency (Xc), and refractive index (Nc). Following previous works,\cite{kuenneth2021polymer, xu2023transpolymer} we assessed the five-fold average $R^2$ on the unseen validation datasets. Among seven supervised models and three self-supervised models, PolyCL achieves the overall best $\textit{R}^{2}$ and four individual best performances across the seven property datasets. PolyCL has a significant advantage in predictive performances over the second-best model in the ionisation energy (Ei), dielectric constant (EPS) and refractive index (Nc) datasets, by 2.4\%, 2.4\%, 4.3\%, respectively. This performance shows that the chemical and structural information captured in the SMILES representation by our model can be generalised to different types of properties, and help to construct more efficient models for quantitative structure-activity relationships. Therefore, the contrastive learning strategy enables the generation of a more expressive representation.

Compared with supervised learning methods, polymer representations produced by self-supervised learning achieved a higher overall performance (Avg. $R^2$) and robustness across all datasets. Only Polymer Genome (PG) fingerprints\cite{kuenneth2021polymer} can reach comparable performance in given tasks -- specifically, the band gap for a chain (Egb) and in the bulk (Egc). However, this fingerprinting method also shows lower robustness, which is observed considering its prediction on different datasets, for example, crystallisation (Xc). ECFP generally exhibits superior overall performance and robustness compared to PG fingerprints; however, in specific tasks involving prediction, PG fingerprints tend to outperform due to their higher target-specificness. In addition, our implementation of graph neural networks suggests that graph representation remains an efficient way to represent polymers; however, these representations do not show better predictive performance than traditional fingerprinting methods.

As an additional assessment, we also assessed the results of fine-tuning, which means that all parameters in both the pre-trained model and the prediction head are unfrozen and fine-tuned (as shown in Fig. S3) Although fine-tuning is not our focus here, we show that our model achieves competitive results compared with other self-supervised models, including polyBERT and Transpolymer, in this experimental setting.

\subsection{Effect of Augmentation Combinations}

\begin{figure*}[!htbp]
\centering
\begin{subfigure}[t]{\textwidth}
    \caption{Explicit Augmentations}
    \includegraphics[width=\textwidth]{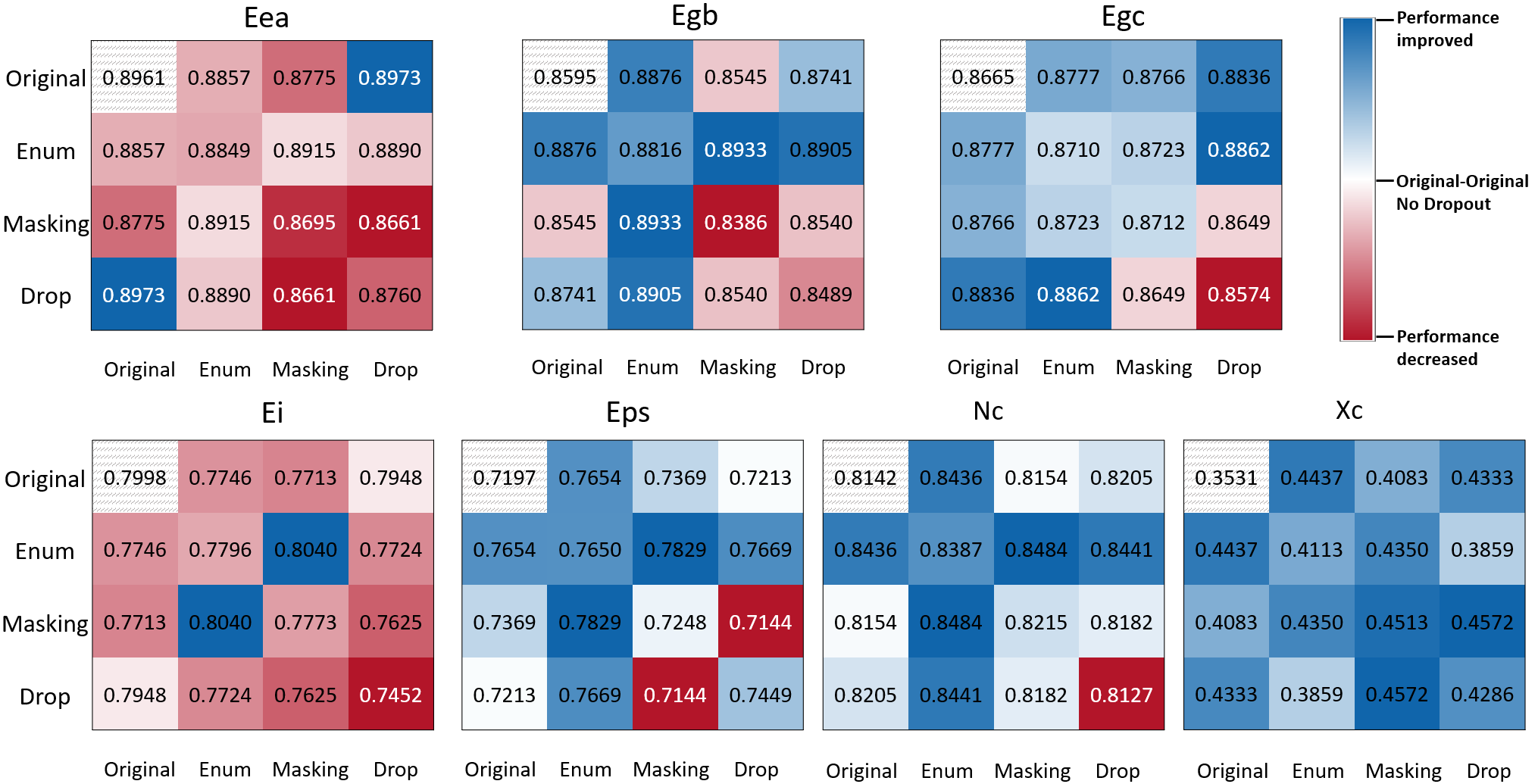}
    \label{fig:explicit}
\end{subfigure}
\begin{subfigure}[t]{0.8\textwidth}
    \caption{Implicit and Mixed Augmentations}
    \includegraphics[width=\textwidth]{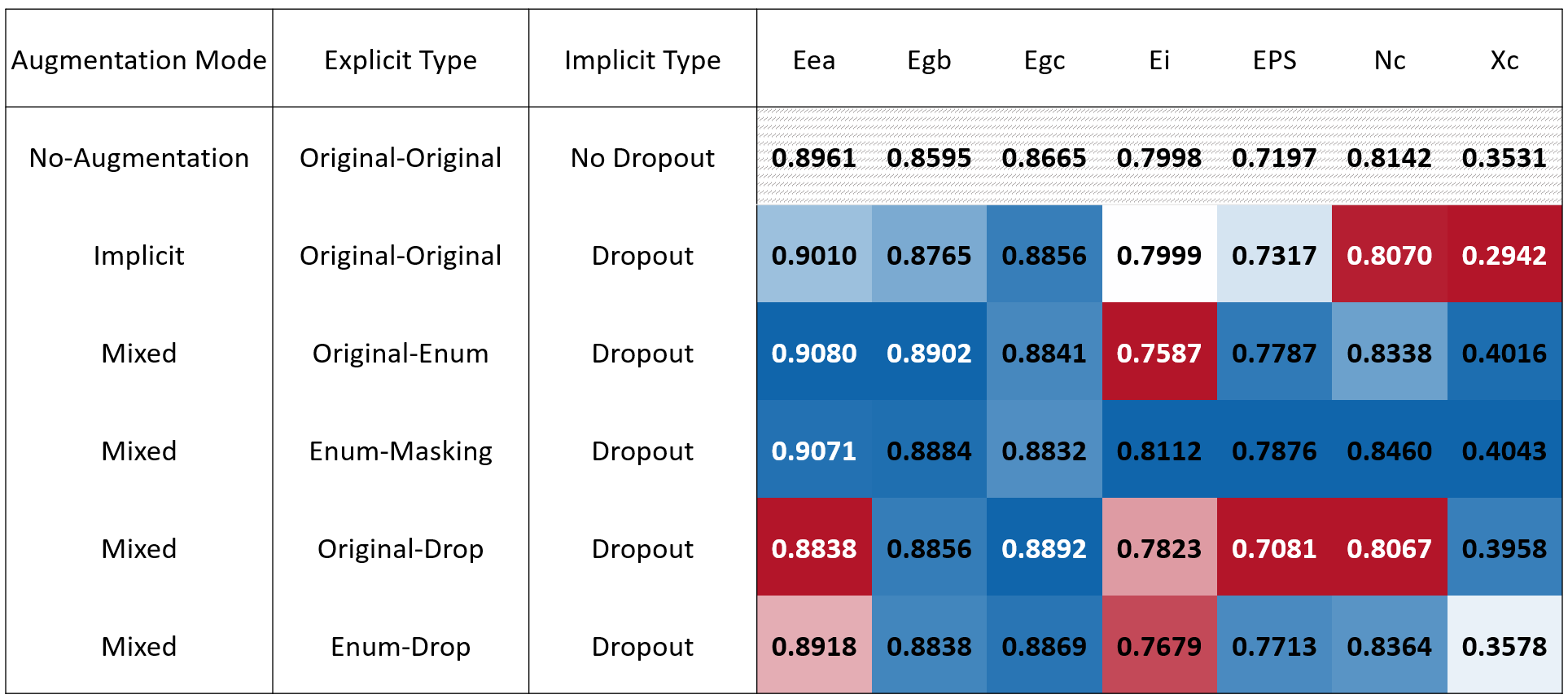}
    \label{fig:mixed_aug}
\end{subfigure}

\caption{Predictive performance of transfer learning evaluated by $R^2$ values on downstream datasets using contrastive learning trained with different augmentation combinations. (a) Explicit augmentations only (where ``Enum'' refers to Enumeration) (b) Implicit and selected mixed augmentation strategy. The striped background cells are the results using the contrastive learning model pretrained with no augmentation (the baseline result). Blue blocks show improved performance relative to the baseline. Red blocks show decreased performance relative to the baseline. The intensity of the colour reflects the magnitude of the deviation.}
\label{fig:augmentations}
\end{figure*}

The combination of augmentation modes can yield differences in the effectiveness of learnt representations. Here, we assess the effect of augmentation combinations by freezing the pre-trained model and only fine-tuning the prediction head during transfer learning, as described in Section \ref{sec:transfer_learning}. The effects of explicit augmentation are shown in Fig. \ref{fig:explicit}, and the effects of implicit and mixed augmentations in Fig. \ref{fig:mixed_aug}. Here, the original input (without augmentations) is used in both branches ({\em{i.e.}} $\bm{x_i}, \bm{x_i} = \bm{x}$), and serves as the contrastive learning baseline (white blocks in Fig. \ref{fig:augmentations}). As seen in Fig. \ref{fig:explicit}, augmentation strategies directly impact the contrastive learning performance. Over the majority of the datasets, augmentations result in enhanced performance compared with the no-augmentation baseline (labelled `Original-Original'). This is especially apparent for Xc, which exhibits low baseline task performance. Here, any combination of augmentations results in a drastic increase in the quality of the representation; this is reflected by the improved performance over the baseline for all augmentation strategies. However, not all combinations of augmentations are suitable for a specific task and the best combination is task-dependent, aligning with the conclusion of previous studies.\cite{tian2020makes}

In addition, explicit augmentations yielded decreased performance relative to the baseline for the electron affinity (Eea) and ionisation energy (Ei) datasets; this is likely due to the already good performance of the baseline model. Downstream tasks exhibiting decreased performance upon inclusion of explicit augmentation may indicate that the property is more closely correlated with structure. Hence explicit augmentations have detrimental effects, which involves making direct, observable changes to the molecule (\textit{e.g.} removing an atom, or breaking a bond, \textit{etc.}). Alternatively, this may also be an indication of the underlying difficulty of the downstream task. For example, Xc is a non-trivial polymer property to assess experimentally and computationally,\cite{venkatram2020predicting} whereas methods to assess electron affinity and ionisation potential are better established.

Considering all of the downstream datasets, we observe that some explicit augmentation combinations demonstrate superior performance relative to others, (Fig. \ref{fig:explicit}). The combinations of Original-Drop and Enumeration-Masking are the best explicit combinations, leading to improved performances compared to the no-augmentation baseline in six of the seven downstream datasets. The second best combination of augmentations are Original-Enumeration and Enumeration-Drop, which demonstrated improved performance for five of the seven downstream datasets. From the results for the augmentation combination study, we observe that including either the original or enumeration augmentation strategies improves performance. It can be intuitively explained why these two augmentations preserve the original and complete semantics of polymer molecules. During Masking and Drop, though the local data structure of polymer-SMILES is preserved, these augmentation types introduce semantic impairment. Therefore, combinations that result in superior performance preserve the full semantics in one branch; this serves as an anchor to give a hint to the parallel branch to complete its full semantics. The strategy behind these combinations might encourage the contrastive pre-training objective to learn more effective representations.

Implicit augmentations (\ref{fig:mixed_aug}) have a unique advantage in creating high-performing contrastive learning strategies, as it outperforms other strategies relative to the baseline for five of the seven datasets; this is comparable to the high-performing explicit augmentation combinations. After confirming the effectiveness of implicit augmentations, we combined the best-performing explicit combinations with implicit augmentations (listed as mixed augmentations in Fig. \ref{fig:mixed_aug}) to identify whether this resulted in improved performance. The addition of implicit augmentations led to varying effects on the performance of explicit combinations. For Original-Enumeration and Enumeration-Masking, implicit augmentations further improved the expressiveness of the resulting representations. However, Original-Drop and Enumeration-Drop suffer from the loss of efficacy. 

Surprisingly, Enumeration-Masking with implicit dropout was the overall best performing combination. This result might demonstrate that the diversified use of augmentation modes is beneficial to the construction of the contrastive learning objective. We can intuitively explain why this combination works. As analysed above, Masking of the original SMILES in one branch conceals part of the information and Enumeration in another branch assists recovery of the original SMILES from its enumerated form. In addition, the semantics in both branches are further disturbed to create slight differences by dropout noises to encourage the comparison. The entire process is comprehensive and effective. Therefore, we chose to apply this augmentation mode, which is the product of the combination of explicit and implicit augmentations, to train our final PolyCL model.

\subsection{Alignment and Uniformity Analysis}

\begin{figure}[H]
    \centering
\includegraphics[width=0.8\linewidth]{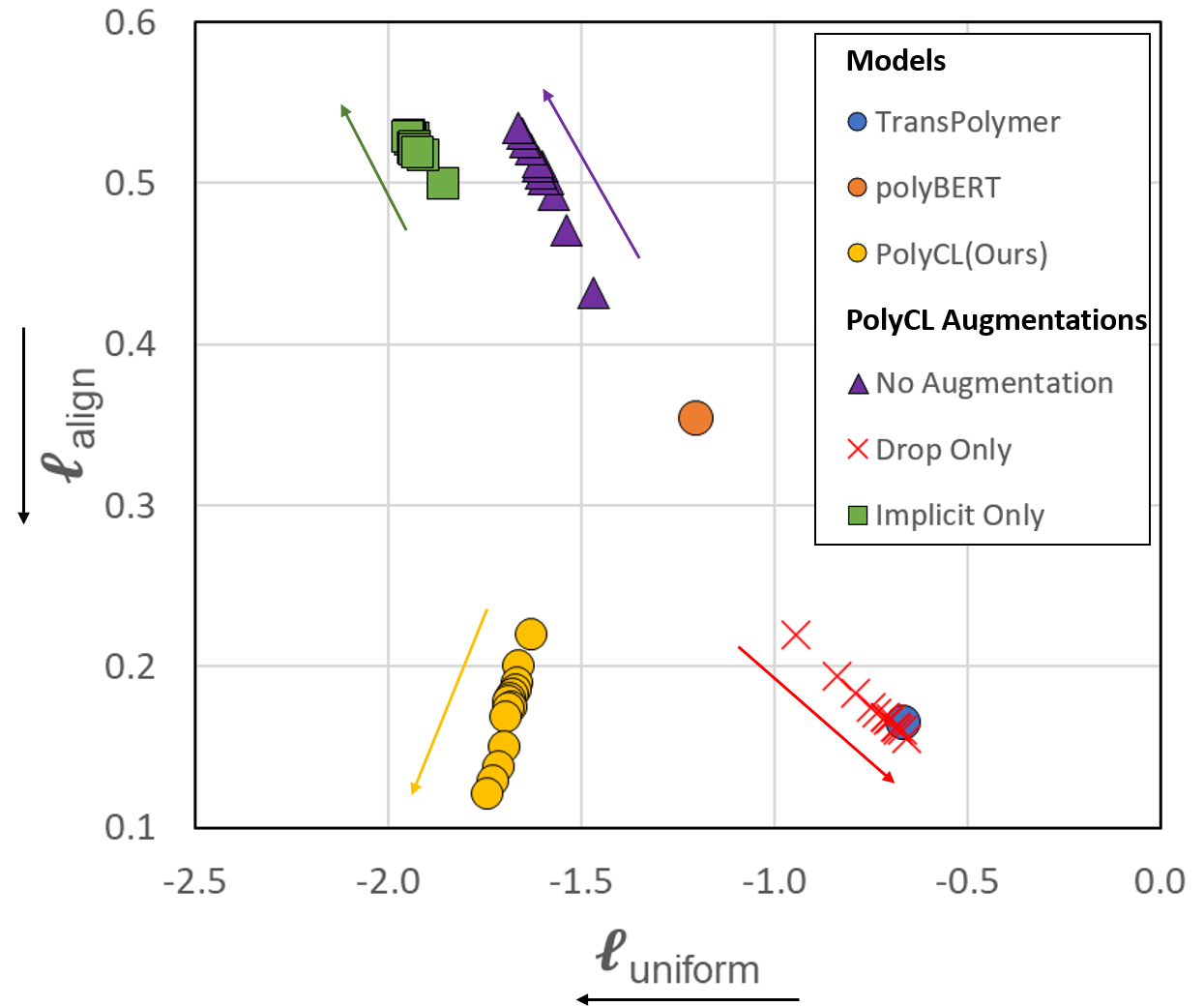}
	\caption{Cross-model comparison on the alignment-uniformity space. For PolyBERT and Transpolymer, the alignment and uniformity of only the final published model is shown. For PolyCL and PolyCL with different augmentation combinations, the intermediate progress during contrastive pre-training is recorded and evaluated with alignment and uniformity. The coloured arrows denote the direction of change during training. The axis label arrows denote the favourable direction.}
	\label{fig:ali_uni}
\end{figure}

As shown in Fig. \ref{fig:ali_uni}, different augmentations yield different training directions in the alignment and uniformity space from the training start point (that of polyBERT). We traced the change of alignment and uniformity during the contrastive pre-training process. In the initial 20\% of total epochs, alignment and uniformity loss was measured at every 2\% checkpoint of total epochs. After that, alignment and uniformity loss was measured at every 20\% of total epochs. In all PolyCL training processes, we observe that the changes in the alignment and uniformity loss of the first 20\% epochs are faster than the remaining epochs, especially in the change of alignment. For the No Augmentation training process (use of only original molecules in both branches), pre-training leads to increased alignment loss but decreased uniformity loss. Since two polymer representations in each positive pair are identical, comparing them is ineffective; this results in the contrastive objective failing to direct the learning of underlying structures by constructing effective positive pairs. On the contrary, the application of only implicit dropout leads to improved distribution (lower uniformity loss) relative to the na\"ive case. However, the magnitude of the alignment loss is comparable and the increase in alignment loss is accompanied by the decrease in uniformity loss. However, the overall change in both metrics is insignificant compared with other augmentation combinations, which indicates that Implicit Only may only have a slight effect on learning representations. The Drop-only case (Drop is applied to both branches) reveals decreased performance, as shown by the high uniformity loss and low alignment loss; this indicates that Drop can still recognise the similarity in feature embeddings, however it fails to capture the diversity in the data. This is further reinforced by the transfer learning results in Fig. \ref{fig:explicit}, where Drop-only only performs well for two of the seven datasets.

Contrary to the other augmentation combinations in Fig. \ref{fig:ali_uni}, PolyCL applies Enumeration-Masking with implicit dropout. From Fig. \ref{fig:ali_uni}, it can be seen that the alignment and uniformity converge to the ideal quadrant ($\ell_{\text {uniform}}$ = -1.7431, $\ell_{\text {align}}$ = 0.1209) during the pre-training guided by the contrastive learning objective -- indicating superior performance. This observation aligns with the transfer learning results in Section \ref{sec:transfer_learning_results}, and with conclusions from previous studies,\cite{wang2020understanding, chen2020simple} which showed that improved alignment and uniformity is generally linked to improved performance of the pre-trained representation. 

We have also evaluated the alignment and uniformity of the pre-trained models polyBERT and Transpolymer. polyBERT ($\ell_{\text {uniform}}$ = -1.1983, $\ell_{\text {align}}$ = 0.3538) has a balanced alignment and uniformity, with both values lying in the middle region, compared to other results. For Transpolymer ($\ell_{\text {uniform}}$ = -0.6640, $\ell_{\text {align}}$ = 0.1649), the alignment loss is comparable to the best contrastive learning models, while the uniformity loss is similar to the Drop-only model.

While PolyCL outperforms other pre-trained models under these two evaluations, it should be noted that polyBERT is the prior of PolyCL. Therefore, the properly trained contrastive pre-training results in the improvement of the model in both alignment and uniformity. It also emphasises the importance of augmentation strategy, as not all augmentations will result in the improvement of both metrics through the training process. Though there is no evident link between the transfer learning performance on specific tasks and the alignment and uniformity, the overall transfer learning performance can be positively correlated to the alignment and uniformity matrix.

\subsection{Representation Space Analysis}

\begin{figure}[p]
\centering
 \includegraphics[width=.48\linewidth]{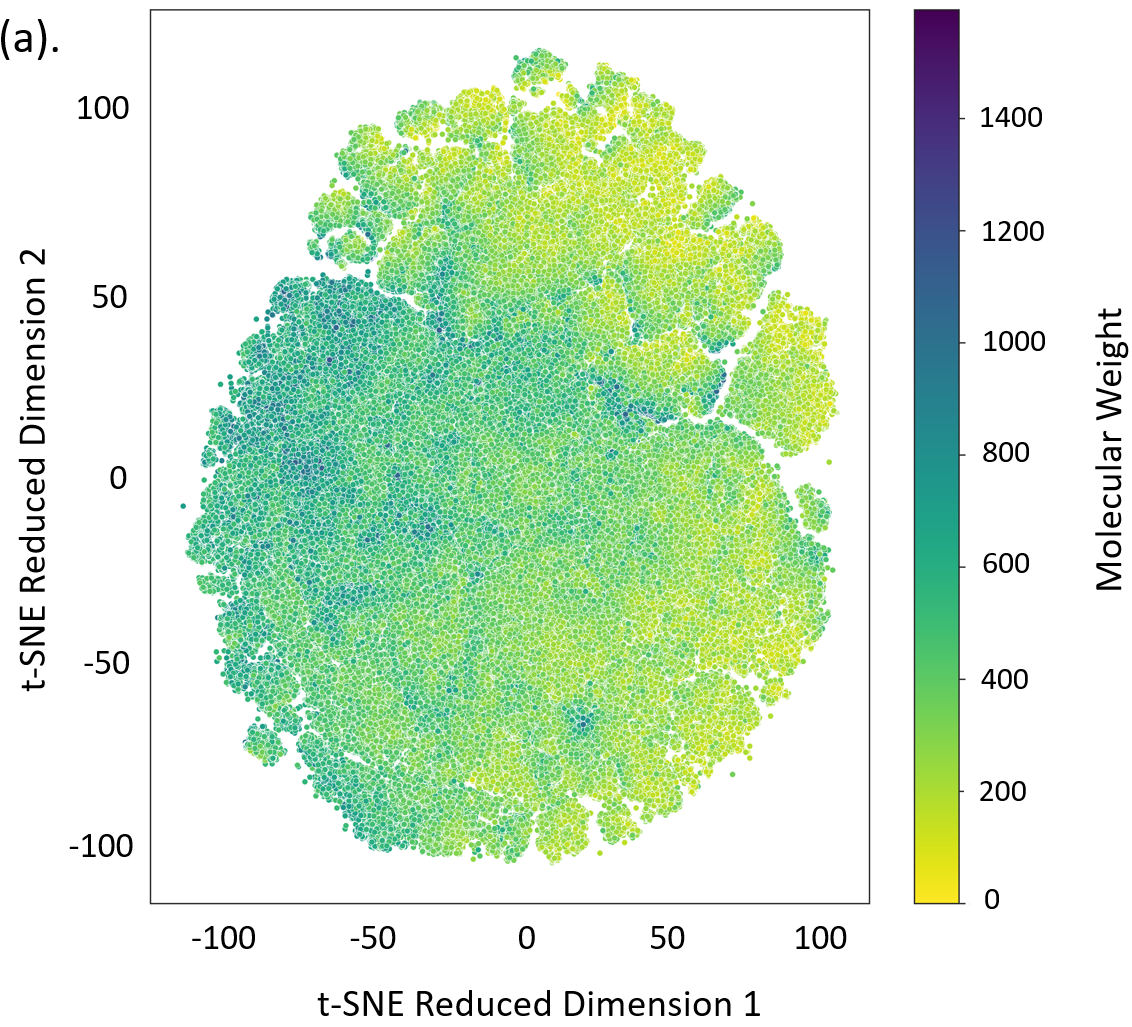}
 \label{fig:tsne_mw}
 \hfill
 \includegraphics[width=.49\linewidth]{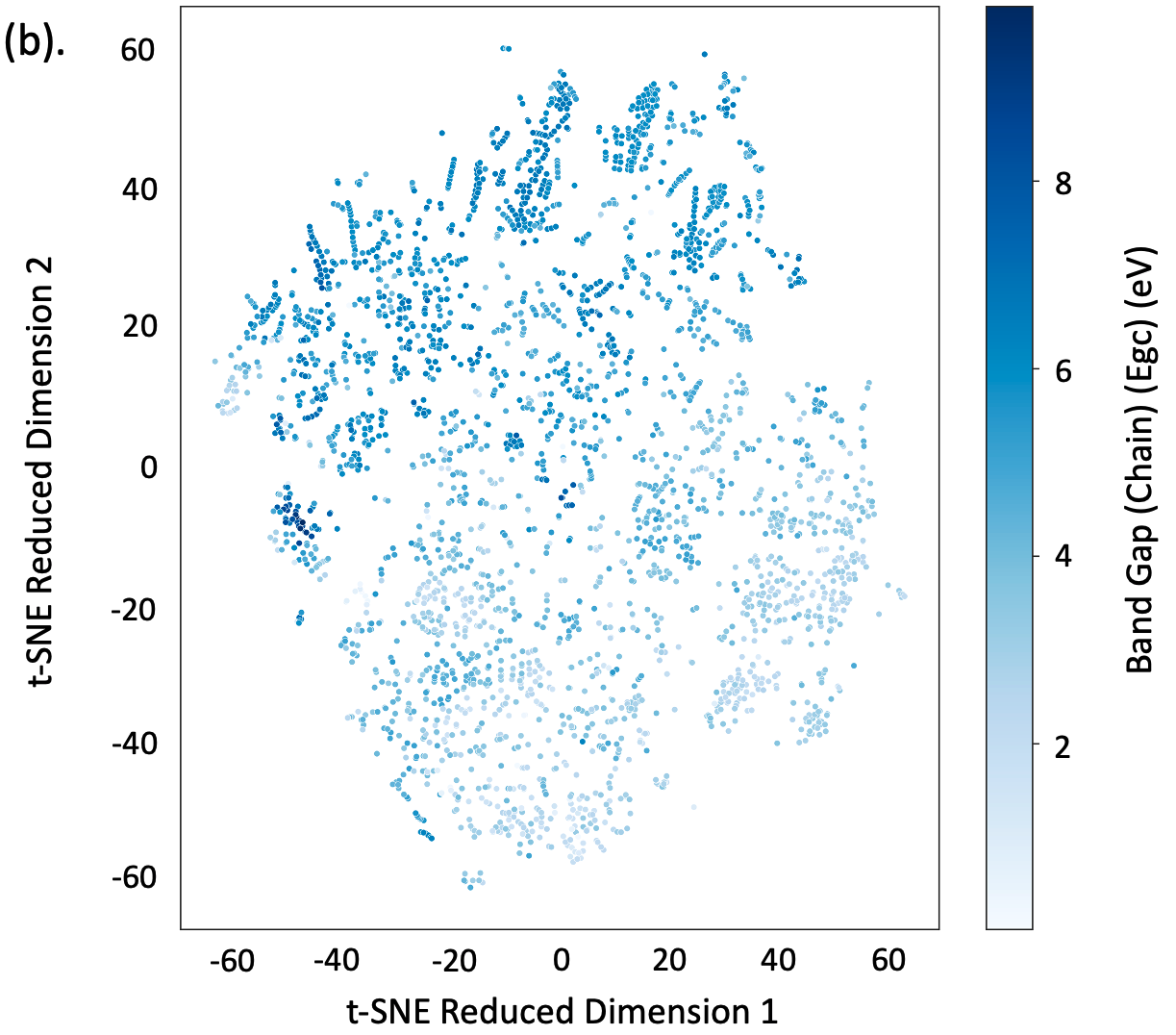}
 \label{fig:tsne_egc}\\

 \includegraphics[width=.78\linewidth]{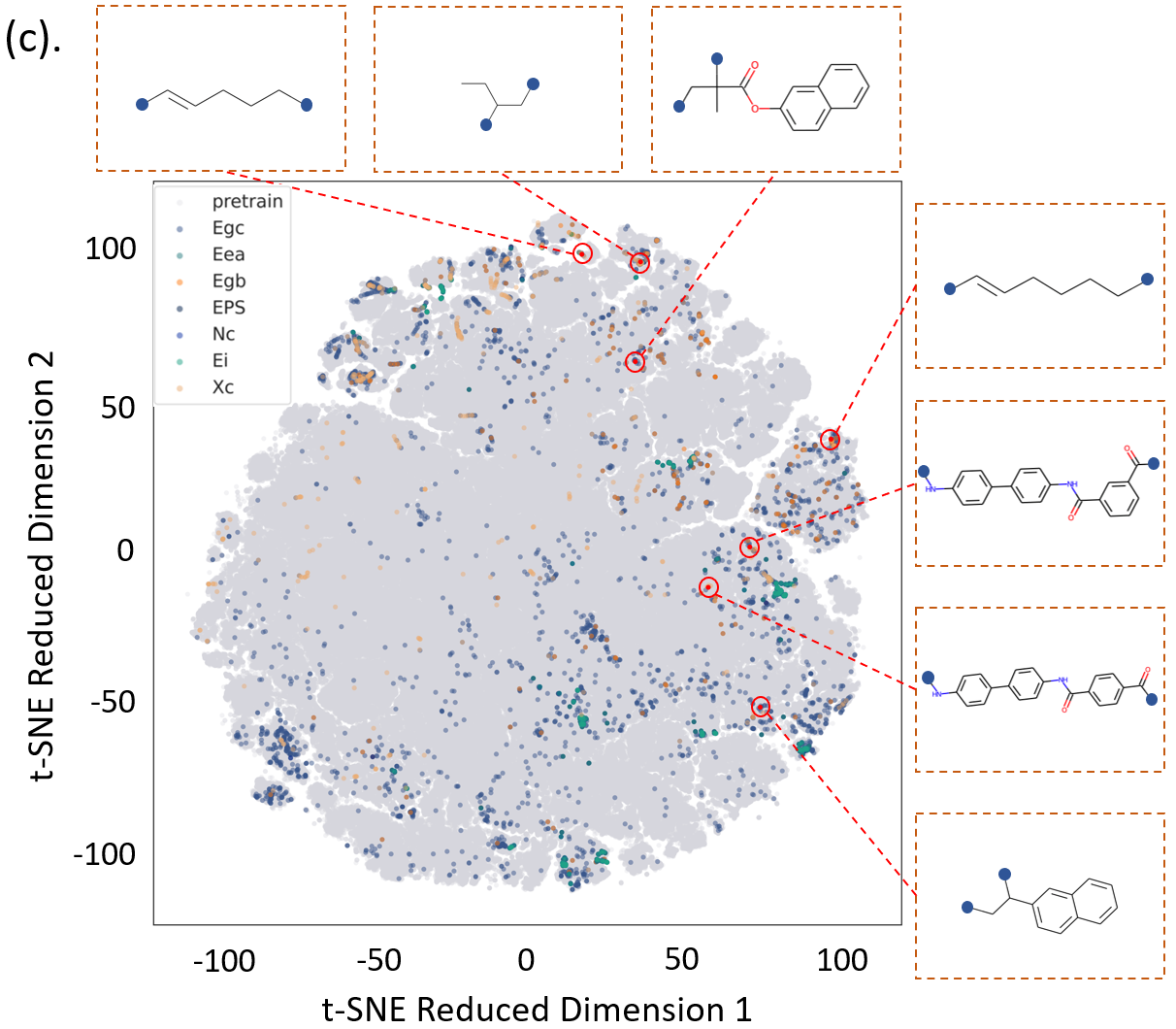} 
 \label{fig:tsne_molvis} \hfill

 \caption{t-SNE dimensional reduction analysis of the polymer representation space learnt by PolyCL. Visualisation of the continuous representation of polymer repeating units: (a) The unsupervised pretrained dataset coloured by molecular weight; (b) The Egc dataset coloured by the band gap (chain) property and (c) all available datasets coloured by the data origin, with selected polymers shown. The blue dot denotes the connection point of the repeating unit to the polymer chain.}
\label{fig:tsne}
\end{figure}

Polymers are transformed into dense and continuous representations by the pre-trained PolyCL. The representation space was evaluated by t-SNE analysis,\cite{van2008visualizing} as shown in Fig. \ref{fig:tsne}. t-SNE analysis arranges data so that points with similar features are plotted in close proximity to each other. Therefore, this method is well-suited to inspecting whether our pre-training method can effectively capture patterns in the learned representations. In Fig. \ref{fig:tsne}(a), the unsupervised dataset was embedded in the representation space and coloured by the molecular weight of each polymer repeating unit. The results show smooth transitions between regions of low to high molecular weight. This suggests that the embedding captures the underlying structure and size of different polymers that correlates closely with their molecular weight differences.

In Fig. \ref{fig:tsne}(b), polymers from a sampled downstream property dataset (Egc) were embedded in the representation space coloured by the value of the band gap (chain) ground truth; the gradient of this representation shows that the sampled downstream property, the chain band gap (Egc), is highly related to the embedded structural features of polymers. In other property datasets, this gradient was also observed (as shown in Fig. S1). Due to the limited number of datapoints in each remaining dataset, the gradient is less evident than the TSNE visualisation of Egc dataset (shown in Fig. \ref{fig:tsne}(b)). 

Our results also suggest that the representation space effectively captures changes in key physical properties implied by the structural features that the original t-SNE was trained on. In Fig. \ref{fig:tsne}(c), all available data is encoded to a representation space and colored by the data source. Here, we observe that the initial, unsupervised dataset comprehensively covers the chemical space encompassed by all of the downstream datasets. We also visualised the molecular structures corresponding to randomly selected points in the embedding. The results show that the structural features learnt by contrastive learning align with human understanding, yet slight divergence. It can be seen from the visualisation that neighbouring representations do not necessarily have similar structures in their molecular graphs. This discrepancy may be due to the different emphasis of SMILES strings and molecular graphs on encoding molecular structures and the special focus of contrastive learning strategies to learn the representations.

\section{Conclusion}

We present a self-supervised pre-training paradigm, PolyCL, that uses contrastive learning to achieve effective polymer representation learning using unsupervised data. We have comprehensively explored varying explicit and implicit augmentation modes and found that the inclusion of both types of augmentations can result in high-performing contrastive learning. Our analysis suggests that the PolyCL-learnt representation excels in preserving chemical information and enhancing model generalisability -- as shown by its superior performance in transfer learning objectives across all seven chemical properties including band gap (both chain (Egc) and bulk (Egb)), electron affinity (Eea), ionisation energy (Ei), DFT-calculated dielectric constant (EPS), crystallisation tendency (Xc), and refractive index (Nc). Additionally, PolyCL exhibits improved chemical property prediction accuracy and robustness across diverse datasets. PolyCL produces high-quality machine-learnt representations, which we expect will be beneficial for a wide range of downstream property-prediction tasks for polymer informatics. The dataset and model are available at: \url{http://github.com/JiajunZhou96/PolyCL}.

\backmatter

\section*{Conflict of Interest}
There are no conflicts of interest to declare.

\section*{Author contributions}
J. Z. developed the PolyCL models and analysed the results. J. Z and Y. Y performed the calculations. A. M. assisted in project design and execution. K. E. J. supervised the project. J. Z. wrote the manuscript and all authors contributed to the final version.

\bmhead{Acknowledgments}

This project made use of time on Tier 2 HPC facility JADE2, funded by EPSRC (EP/T022205/1). A. M. M is supported by the Eric and Wendy Schmidt AI in Science Postdoctoral Fellowship, a Schmidt Sciences program. K. E. J acknowledges the European Research Council through Agreement No. 758370 (ERC-StG-PE5-CoMMaD) and the Royal Society for a University Research Fellowship.

\bibliography{sn-bibliography}%



\section*{Supplementary material}

# Supporting Information:
# PolyCL: Contrastive Learning for Polymer Representation Learning via Explicit and Implicit Augmentations

Jiajun Zhou,[†] Yijie Yang,[†] Austin M. Mroz,[†,‡] and Kim E. Jelfs[*,†]

†Department of Chemistry, Molecular Sciences Research Hub, Imperial College London, White City Campus, Wood Lane, London, W12 0BZ, U.K.

‡I-X Centre for AI in Science, Imperial College London, White City Campus, Wood Lane, London, W12 0BZ, U.K.

E-mail: k.jelfs@imperial.ac.uk

# Contents

# S1 Downstream Datasets

Table S1: Downstream datasets sourced from Xu *et al.*[S1]

| Dataset | Property | Size | Unit |
|---------|----------|------|------|
| Egc | bandgap(chain) | 3380 | eV |
| Egb | bandgap(bulk) | 561 | eV |
| Eea | electron affinity | 368 | eV |
| Ei | ionisation energy | 370 | eV |
| Xc | crystallisation tendency | 432 | % |
| EPS | dielectric constant | 382 | - |
| Nc | refractive index | 382 | - |

# S2 Transfer Learning Performance

Here, we report the average root mean square error of the transfer learning task, where only the prediction head was fine-tuned and the pre-training model was frozen.

Table S2: The average root mean square error on the unseen validation datasets with Five-Fold Cross-Validation. #Params is the parameter count in the model.

| Model information | | Datasets | | | | | | |
|---|---|---|---|---|---|---|---|---|
| Model | #Params | Eea | Egb | Egc | Ei | EPS | Nc | Xc |
| $RF_{ECFP}$ | - | 0.4295 | 0.7163 | 0.5623 | 0.4978 | 0.6220 | 0.1181 | 17.7347 |
| $XGB_{ECFP}$ | - | 0.4342 | 0.7184 | 0.5674 | 0.5151 | 0.6336 | 0.1173 | 18.5263 |
| $NN_{ECFP}$ | 264K | 0.4095 | 0.6989 | 0.5324 | 0.4835 | 0.5543 | 0.1046 | 18.2696 |
| $GP_{PG}$[S2] | - | 0.32 | 0.55 | 0.48 | 0.42 | 0.53 | 0.10 | 24.42 |
| $NN_{PG}$[S2] | - | 0.32 | 0.57 | 0.49 | 0.45 | 0.54 | 0.10 | 20.74 |
| GCN | 70K | 0.4146 | 0.8605 | 0.7009 | 0.5040 | 0.6472 | 0.1655 | 19.3743 |
| GIN | 218K | 0.3717 | 0.7854 | 0.6663 | 0.4590 | 0.6197 | 0.1456 | 18.5009 |
| TransPolymer[S1] | 82.1M | 0.3446 | 0.6266 | 0.5498 | 0.4450 | 0.5475 | 0.1033 | 17.4086 |
| PolyBERT[S3] | 25.2M | 0.3272 | 0.6636 | 0.5448 | 0.4672 | 0.5326 | 0.1063 | 17.6989 |
| PolyCL | 25.2M | 0.3265 | 0.6458 | 0.5329 | 0.4180 | 0.5111 | 0.0933 | 18.1747 |

# S3 Fine-Tune Performance of Self-Supervised Learning Models

Here, the average $R^2$ values and root mean square errors of the fine-tuning task are shown, where both the pre-trained model and prediction head were fine-tuned.

Table S3: The average of $R^2$ on the unseen validation datasets with Five-Fold Cross-Validation on self-supervised models. #Params is the parameter count in the model. The boldings indicate the best results.

| Model information | | Datasets | | | | | | | |
|---|---|---|---|---|---|---|---|---|---|
| Model | #Params | Eea | Egb | Egc | Ei | EPS | Nc | Xc | Avg.$R^2$ |
| TransPolymer[S1] | 82.1M | 0.9179 | **0.9365** | 0.9177 | 0.8278 | 0.7966 | 0.8674 | 0.4500 | 0.8163 |
| PolyBERT[S3] | 25.2M | **0.9365** | 0.9243 | 0.9163 | 0.8359 | **0.8081** | 0.8653 | 0.4428 | **0.8185** |
| PolyCL | 25.2M | 0.9317 | 0.9273 | **0.9186** | **0.8491** | 0.8038 | **0.8699** | 0.4250 | 0.8179 |

Table S4: The average root mean square error on the unseen validation datasets with Five-Fold Cross-Validation. #Params is the parameter count in the model.

| Model information | | Datasets | | | | | | |
|---|---|---|---|---|---|---|---|---|
| Model | #Params | Eea | Egb | Egc | Ei | EPS | Nc | Xc |
| TransPolymer[S1] | 82.1M | 0.3031 | 0.4901 | 0.4480 | 0.4052 | 0.5014 | 0.0859 | 17.3788 |
| PolyBERT[S3] | 25.2M | 0.2659 | 0.5304 | 0.4517 | 0.3972 | 0.4850 | 0.0875 | 17.6082 |
| PolyCL | 25.2M | 0.2757 | 0.5243 | 0.4457 | 0.3799 | 0.4923 | 0.0858 | 17.8229 |

# S4  Pre-trained Model Comparison

Table S5: Comparison of pre-trained Polymer Foundation Models

| Model | Training Set Size | Training Strategy | #Params |
|---|---|---|---|
| TransPolymer | 5M | MLM | 82.1M |
| PolyBERT | 80M | MLM | 25.2M |
| PolyCL | 80M + 1M[1] | Contrastive Learning | 25.2M |

[1] PolyCL has PolyBERT as its pre-trained prior. PolyCL is trained with 1M datapoints. MLM indicates masked language modelling. #Params is the parameter count for each model.

# S5 Results of TSNE on Downstream Datasets

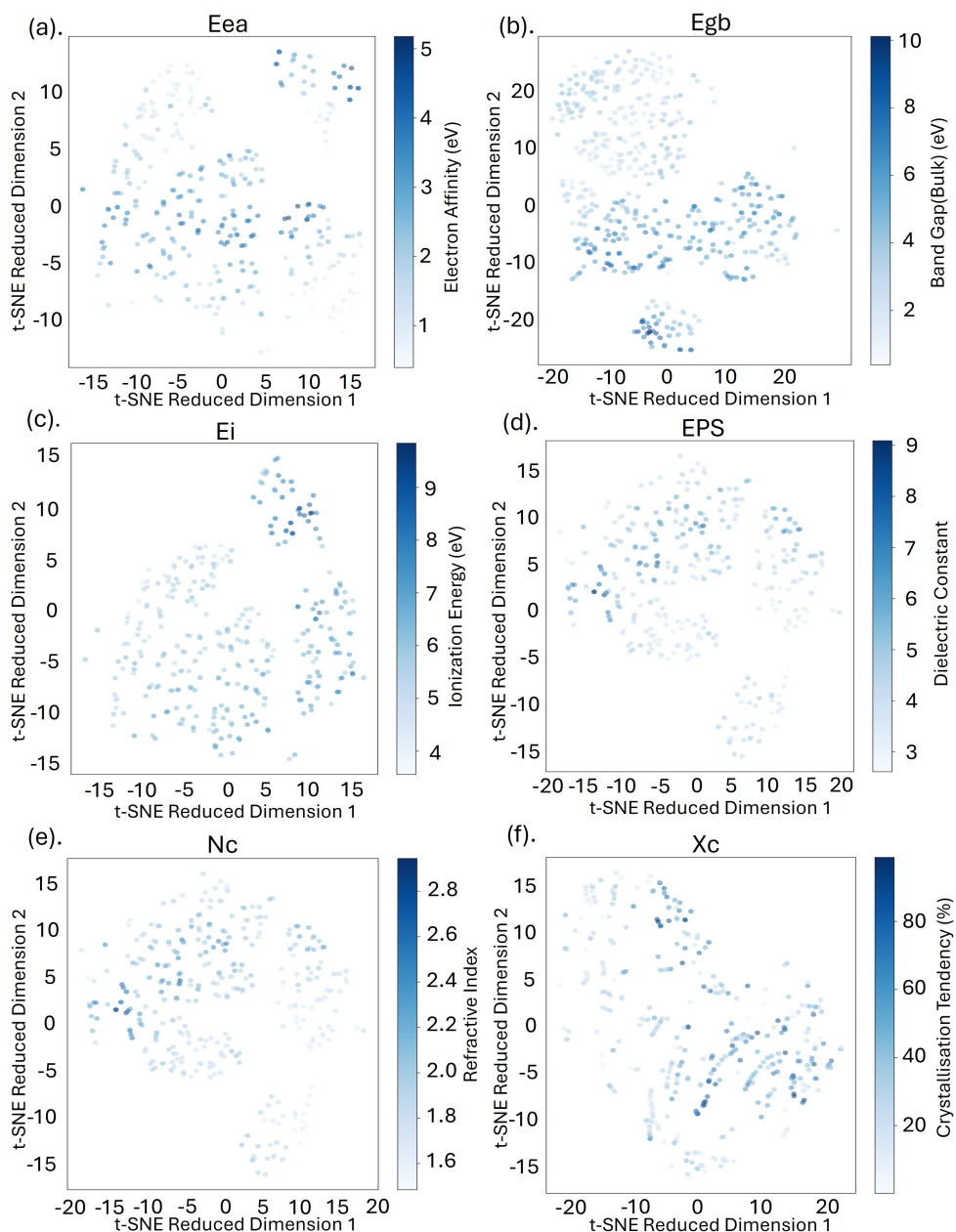

Figure S1: t-SNE dimensional reduction analysis of the polymer representation space learnt by PolyCL. Visualisation of the continuous representation of polymer repeating units. Each figure on the downstream dataset is coloured by the corresponding value of the property in the downstream dataset. (a) Eea dataset coloured by electron affinity. (b) Egb dataset coloured by bandgap (bulk). (c) Ei dataset coloured by ionisation energy (d) EPS dataset coloured by dielectric constant (e) Nc dataset coloured by refractive index (f) Xc dataset coloured by crystallisation tendency

# S6 Training Details of Supervised Models

A five-fold cross-validation process was applied to each algorithm. The fold average of all $R^2$ and $RMSE$ values on each unseen validation set were taken to evaluate the model performance.

## S6.1 Random Forest with ECFP

We used RDKit[S4] to transfer the polymer-SMILES to 512-bit ECFP fingerprints. The scikit-learn package[S5] was used to construct the random forest regressor and hyperparameters were tuned manually. The final random forest model we used is:

`sklearn.ensemble.RandomForestRegressor`
`(bootstrap=False,max_features='sqrt',random_state=72)`

where all other parameters in the above implementation remain the default parameters.

## S6.2 XGBoost with ECFP

We constructed 512-bit ECFP fingerprints as shown in Section S6.1. The XGBoost[S6] package was used to construct our xgboost model. After tuning, the final model is:

`xgboost.XGBRegressor`
`(max_depth=6,min_child_weight=3,colsample_bytree=0.8,random_state=72)`

where all other parameters in the above implementation remain the default parameters.

## S6.3 Neural Network with ECFP

We obtained the same ECFP fingerprints as shown in Section S6.1. We employed a simple two-layered MLP with a dropout ratio of 0.1. As an early stopping mechanism, a patience value of 100 was used with the epoch size set to a large value. During training, the batch size was set to 16 and the learning rate was 0.001.

## S6.4 Gaussian Process and Neural Network with Polymer Genome

The performances are directly taken from the originally reported results by Kuenneth *et al.*[S2]

## S6.5 Graph Convolutional Network

We abstracted polymers to two-dimensional undirected molecular graphs. We follow the featurisation method used by Hu *et al.*[S7] to encode nodes and edges to the graph representation using feature sets. Features are extracted using RDKit.[S4] The connecting point in polymer data is marked as a special node indexed by 0 in the node set.

Table S6: Feature sets for nodes and edges in molecular graphs.

| Component | Feature | Feature Set Details |
|:---:|:---:|:---:|
| Node | Atomic number | [0, 119] |
| Node | Chirality | unspecified, tetrahedral CW, tetrahedral CCW, other |
| Edge | Bond type | single, double, triple, aromatic |
| Edge | Bond direction | none, end-upright, end-downright |

The nodes and edges were integrated by using PyTorch Geometric[S8] to construct molecular graphs.

A molecular graph $G = (V, E)$ is constructed by collections of nodes $V$ and edges $V$, where the node features are denoted by $X_v$ for $v \in V$. GNN layers learn to create embeddings $h_v$ at the node level, by aggregating the features $X_v$ from neighbouring nodes. Consequently, the node embedding after $k$-th layers encompasses the information within its $k$-hop neighbourhood. The $k$-th layer of a GNN can be described by:

$$\begin{aligned} a_v^{(k)} &= \text{AGGREGATE}^{(k)}\left(\left\{h_u^{(k-1)} : u \in \mathcal{N}(v)\right\}\right) \\ h_v^{(k)} &= \text{COMBINE}^{(k)}\left(h_v^{(k-1)}, a_v^{(k)}\right) \end{aligned} \quad (1)$$

In practice here, we used two different architectures of $\text{AGGREGATE}^{(k)}$ to construct GNNs to process polymer graphs. In graph convolutional networks (GCN), element-wise

mean pooling is proposed.[S9] Therefore, the update rule can be summarised as:

$$h_v^{(k)} = \text{ReLU}\left(W^{(k)} \cdot \text{MEAN}\left(\{h_u^{(k-1)} : u \in \mathcal{N}(v) \cup \{v\}\}\right)\right) \quad (2)$$

Alternatively, graph isomorphism networks (GIN)[S10] use a sum aggregation function and non-linear transformation by an MLP:

$$h_v^{(k)} = \text{MLP}^{(k)}\left(\left(1 + \epsilon^{(k)}\right) \cdot h_v^{(k-1)} + \sum_{u \in \mathcal{N}(v)} h_u^{(k-1)}\right) \quad (3)$$

For the final layer node representation $h_v^{(k)}$, the READOUT function is used to aggregate node features to obtain the graph representation:

$$h_G = READOUT(h_v^{(k)}|v \in G) \quad (4)$$

For both types of graph neural networks, GCN and GIN, we used the same set of hyperparameters except for the aggregation process inherently defined differently as shown in Equation 2 and 3. We used 3 layered GNN with a dropout rate of 0.1. The node and edge attributes were embedded in 128-dimensional space. The supervised training process was set to last for 100 epochs. An early stopping mechanism was implemented with a patience value of 50 epochs.



\end{document}